\documentclass[a4paper]{article}
\usepackage{INTERSPEECH2021}

\usepackage{arydshln}
\usepackage{graphicx}
\usepackage{subcaption}
\usepackage{tikz}
\usepackage{multicol}
\usepackage{multirow}
\usepackage{xcolor}

\title{Understanding Medical Conversations: \\
Rich Transcription, Confidence Scores \& Information Extraction}
\name{Hagen Soltau, Mingqiu Wang, Izhak Shafran, Laurent El Shafey}
\address{Google}
\email{\{soltau, mingqiuwang, izhak, shafey\}@google.com}
\date{January 2021}

\begin{document}
\maketitle

\begin{abstract}
In this paper, we describe novel components for extracting clinically relevant information from medical conversations which will be available as Google APIs. We describe a transformer-based Recurrent Neural Network Transducer (RNN-T) model tailored for long-form audio, which can produce rich transcriptions including speaker segmentation, speaker role labeling, punctuation and capitalization. On a representative test set, we compare performance of RNN-T models with different encoders, units and streaming constraints. Our transformer-based streaming model performs at about 20\% WER on the ASR task, 6\% WDER on the diarization task, 43\% SER on periods, 52\% SER on commas, 43\% SER on question marks and 30\% SER on capitalization. Our recognizer is paired with a  confidence model that utilizes both acoustic and lexical features from the recognizer. The model performs at about 0.37 NCE. Finally, we describe a RNN-T based tagging model. The performance of the model depends on the ontologies, with F-scores of 0.90 for medications, 0.76 for symptoms, 0.75 for conditions, 0.76 for diagnosis, and 0.61 for treatments. While there is still room for improvement, our results suggest that these models are sufficiently accurate for practical applications. 
\end{abstract}

\section{Introduction}
\label{sec:intro}
Medical conversations are an important source of information for clinical documentation and decision making. As the automatic speech recognition (ASR) and natural language processing (NLP) technology has matured, there has been considerable research interest in automatically extracting useful information from these conversations~\cite{Enarvi2020,Wang2020b,Mani2020}. Previous work has focused on certain sub-problems such as summarization from transcripts, error analysis of ASR, or adapting tools from text to ASR transcripts. Instead, we focus on the minimal components necessary for extracting clinical information from raw audio, starting with a rich transcription of the medical conversation, as described in Section~\ref{sec:rich_trans}. We delve into estimation of confidence scores in Section~\ref{sec:confidence}, which is one of the most requested features in a medical ASR so users can decide when to trust the correctness of the recognized outputs. Finally, we describe a Recurrent Neural Network Transducer (RNN-T) based tagger in Section~\ref{sec:nlp} which is the first of its kind to our knowledge.

\section{Rich Transcription}
\label{sec:rich_trans}
Speech recognition is the first step to understand medical conversations. Before delving into the details of our ASR models, it is useful to understand the nature of medical conversations and the associated challenges. 

The audio in medical conversations is substantially different from applications such as medical dictations or digital assistant, in terms of speakers, speaking styles and duration. The conversations may involve 2-5 speakers -- doctor, nurse, patient, caregiver and other clinical support staff. The audio is typically captured by a far-field microphone, with a large variance in the distance from the speakers. In contrast to other conversational corpora (e.g. Switchboard), the style is collaborative, with relatively low-cross talk, as all parties have vested interest in addressing the issue at hand efficiently and effectively. More importantly, the conversations are fairly long. While the average duration is about 11 minutes, some conversations exceed more than an hour. Note, several recent papers treat long-form audio, for example on LibriSpeech task, as concatenation of short segments, which results in sub-optimal performance. For more details on the corpus, see~\cite{shafey19}.

Since the overall goal is to understand medical conversations, we need to create a {\it rich} transcription of the conversations that include not only punctuation, capitalization and confidence scores but also perform the diarization task that involves inferring speaker turns and associating them with the speakers so we know who spoke what and when. Since our task is akin to the NIST Rich Transcription tasks, we borrowed confidence and diarization metrics from them~\cite{Fiscus2006}. In the following, we discuss how these requirements impact modeling choices.

The factorization of the speech recognition models into separately trained acoustic $P(X|W)$ and language models $P(W)$, first formulated as Hidden Markov Model (HMM), continues to be a dominant choice for ASR. In this framework, there is a lack of strong coupling between acoustic and language models even when the acoustic models are trained with discriminative functions such as MMIE or MPE~\cite{Woodland2002,PoveyGKW03} or the acoustic models are modeled by LSTMs~\cite{Hochreiter1997} or transformers~\cite{Vaswani2017,Wang2020}. This lack of coupling makes it difficult to utilize both acoustic and linguistic cues simultaneously in rich transcription. 

Alternative approaches based on {\em sequence-to-sequence models with an attention mechanism} do jointly optimize the acoustic (encoder) and the language (decoder) components~\cite{Bahdanau2015,Chorowski2016,Chan2016}. While these are apt models for rich transcription, their output suffers from deletions and repetitions, particularly when the input sequence is long. This has been somewhat alleviated in practice by {\it ad hoc} methods such as input-coverage penalties and monotonic alignment constraints. In previous work on medical conversations, the deletion issue went unnoticed because the models were trained and evaluated on selected segmented data where difficult to recognize segments were excluded~\cite{Chiu2018}.

{\em Recurrent Neural Network Transducer (RNN-T) models} are similar to attention-based sequence-to-sequence models in that the encoder and the decoder are optimized jointly with a single cost function~\cite{Graves2012}, making these models also apt for rich transcription. Furthermore, the inference is performed in a time-synchronous fashion with a beam search process that accounts for {\em every} input frame. So, unlike the attention model, they do not suffer from the deletion or duplication problems and they are also well-suited for streaming applications~\cite{Rao2017}. In our previous work, we demonstrated that RNN-Ts can accurately predict speaker turns and two speaker role labels in a non-streaming application using bidirectional LSTM encoders~\cite{shafey19}. In this paper, we investigate how RNN-T models can be applied in a streaming ASR application for rich transcription of long medical conversations with more speakers.

\subsection{Experimental Setup} \label{sec:asr_expts}
Our experiments were performed on a large corpus of about $100$K ($\approx 15$K hours) manually transcribed audio recordings of clinical conversations between physicians and patients. Apart from medical providers (physicians or nurses) and patients, the conversations may include caregivers and other support staff. In total, we model 4 speaker roles, in contrast to~\cite{shafey19}, where we restricted the experiments to 2 speaker roles.

The models are evaluated on whole conversations which are segmented automatically using a neural network based speech detector. Additionally, to understand the impact of model performance on different utterances, we created three versions of test sets with different average segment duration, specifically, $5$, $10$ and $15$ seconds. These segments were created using the Viterbi alignment with the reference transcripts.

\begin{table}[ht]
\centering
\begin{tabular}{|l|c|c|c|}
\hline
Segmentation &  Avg (sec) & Max(sec) & Segments \\ \hline
Viterbi-5     & 7.8  & 389 & 37981 \\
Viterbi-10    & 12.8 & 399 & 23125 \\
Viterbi-15    & 17.8 & 402 & 16644 \\
Automatic    & 38.2 & 60  & 7759 \\
Unsegmented  & 674.1 & 3620 & 440 \\ \hline
\end{tabular}
\caption{Evaluation sets with different duration.}
\label{asr:seg}
\vspace{-.3in}
\end{table}    

\subsection{Choice of Encoder and Units} \label{sec:encoder}
Streaming transcription of long sequences poses a challenge to the encoders. They need to capture sufficient temporal dependencies necessary for accurate prediction without observing the whole input sequence. One additional complexity that exacerbates this challenge is a mismatch between training and testing scenarios. The training data is typically segmented so that they can be trained efficiently in large batches on accelerators. Bidirectional encoders tend to be less sensitive to this mismatch, but auto-regressive encoders have been shown to exhibit high deletion rates when processing long utterances. Several workarounds have been proposed, including resetting the LSTM state every $N$ frames~\cite{Hasim2014}.

Transformers with limited, fixed size, attention have been proposed for streaming applications~\cite{Qian2020}. Instead of computing attention scores over the entire sequence, this modified version of transformer adopts an approach where the computation is limited to a fixed window and the receptive field is build up hierarchically with the number of layers~\cite{Povey2020}. They demonstrated performance gains on Librispeech~\cite{Panayotov2015}, which has relatively shorter segments than our task.

While longer units such as morphemes~\cite{morfessor2}  may be beneficial in bidirectional encoders, the lack of sufficient future (right) context in unidirectional encoders could be more debilitating than smaller units like graphemes. 
The {\em output} frame rate needs to be tied to the size of the target units. This consideration led to tailoring the output rate of graphemes and morphemes to 40ms and 80ms respectively, achieved using 2 and 3 sub-sampling or TDNN layers respectively, similar to our previous work~\cite{soltau2017b}.

\subsection{ASR Performance}
The results are summarized in table~\ref{asr:wer} for streaming and non-streaming models. The key points are: (a) Transformer encoders perform about $20\%$ better than bidirectional LSTM with TDNN layers, regardless of units or sequence lengths; (b) For the non-streaming transformer encoder, the performance does not depend on the choice of units; (c) For the full context LSTM/TDNN encoder, the performance is about $10\%$ better when using larger morpheme units; (d) The loss of right context leads to $10\%$ to $20\%$ degradation in performance, with higher degradation for the larger morphemes; and (e) When using left context only, the shorter grapheme units outperform the morpheme units by about $10\%$ in particular for longer sequences, and consistently across both encoders.

\begin{table}[h]
\centering
\begin{tabular}{|l|l|c|c|c|c|c|} \hline
Enc.   & Units     & Vit-5 & Vit-10 & Vit-15 & Auto & UnSeg \\ \hline
\multicolumn{7}{|c|}{non-streaming} \\ \hline
L & G   & 23.3  & 22.0   & 21.4   & 20.9 & 20.0  \\ \hline
L & M & 22.2  & 20.1   & 19.5   & 19.1 & 18.4  \\ \hline
T & G  & 17.4  & 16.8   & 16.6   & 16.5 & 16.1 \\ \hline
T & M  & 17.8  & 17.0   & 16.8   & 16.5 & 16.2  \\ \hline
\multicolumn{7}{|c|}{streaming} \\ \hline
L & G & 25.5  & 24.7   & 24.4   & 24.8 &  24.4  \\ \hline
L & M & 24.1  & 23.2   & 23.3   & 25.1 &  25.5  \\ \hline
T & G & 20.7  & 20.0   & 19.7   & 20.3 &  19.6 \\ \hline
T & M & 19.8  & 19.0   & 19.5   & 21.9 &  21.2 \\ \hline
\end{tabular}
\caption{WER comparison of encoders and units with and without streaming constraint. Encoder Types: L=LSTM/TDNN, T=Transformer, Units: G=Graphemes, M=Morphemes, Auto: deployment scenario.}
\label{asr:wer}
\vspace{-.2in}
\end{table}

\subsection{Rich Transcription Performance}
The performance of speaker segmentation and role predictions are measured in {\em Word Diarization Error Rate (WDER)}, the percentage of words in the transcript decorated with the right speaker tag, akin to metrics in NIST evaluations~\cite{RT2003}. The punctuation performance is measured using {\em Slot Error Rate (SER)}, as defined in previous work~\cite{Makhoul99performancemeasures,kolar:hal-01843550}. To measure capitalization performance, a special token is used to indicate an upper-case word, and we measure the slot error rate for that token.

\begin{table}[h]
\centering
\begin{tabular}{|l|c|c|c|c|c|c|} \hline
Context & WDER & WER & '.' & ',' & '?' & Cap \\ \hline
NS      & 3.2  & 16.1 & 35.1 & 48.5 & 36.2 & 25.0 \\ \hline
S       & 9.7  & 20.3 & 42.7 & 52.0 & 42.8 & 29.5 \\ \hline
S-ESS & 5.6 & 19.6 & - & - & - & - \\ \hline
\end{tabular}
\caption{Comparison of rich transcription performance under three conditions: non-streaming (NS), streaming (S) and streaming with encoder state saving (S-ESS).}
\label{result:rt}
\vspace{-.1in}
\end{table}

Table~\ref{result:rt} compares rich transcription performance under non-streaming (NS), streaming (S) and streaming with state saving across segments in the encoder (S-ESS). It is worth noting that extending the left context across speech segments is crucial in maintaining high accuracy in speaker role prediction when going from non-streaming (3.2\% WDER in NS) to streaming scenario (5.6\% WDER in S-ESS). Without the state saving mechanism, the performance drops substantially in the streaming case (9.7\% WDER in S).

Our results show that capitalization is an easier task than predicting punctuation and among punctuation periods and question marks are easier than commas. There is a performance loss associated with the streaming scenario which cannot be recovered with state saving (not shown). Unfortunately, there is scant recent work on punctuation and capitalization. Models pre-trained on text corpora have been applied for post-processing ASR output to predict punctuation and capitalization~\cite{sunkara-etal-2020-robust}. They use F1-score which does not take into discounts deletions and insertions by a factor of $2$~\cite{Makhoul99performancemeasures}. In the context of NIST Rich Transcription evaluation, the performance of predicting punctuation was reported at a slot error rate of 68\% and 53.5\% for ASR output and manual transcripts respectively~\cite{kolar:hal-01843550}. Our results compares favorably and moreover our model does not require post-processing ASR output like the above mentioned previous work.

\section{Confidence Models}
\label{sec:confidence}

There has been more than two decades of research on developing a robust and accurate confidence estimate of ASR outputs~\cite{Siu1999,Kemp1997}. In the early days when the acoustic and the language models were estimated separately, the dominant paradigm involved extracting acoustic and linguistic features from the ASR lattices and predicting the scores using a separate classifier~\cite{Kemp1997}. In neural network models, one of the popular techniques involves modifying the posterior probability (softmax output) of the primary recognizer or classifier itself~\cite{Guo2017}. Unfortunately, the RNN-T models have very skewed posterior probabilities compared to, for example, CTC models which do not have an integrated language model~\cite{Graves2006b}. Moreover, the RNN-T models are optimized using negative log-likelihood which can be decreased by increasing the value of logits for correct samples~\cite{muller2019does}. More recently, sequence-to-sequence models with attention has been proposed that has the flexibility to use various features from the ASR model~\cite{li2020confidence, qiu2021learning}.

\begin{figure}[h]
    \centering
    \includegraphics[width=0.95\columnwidth]{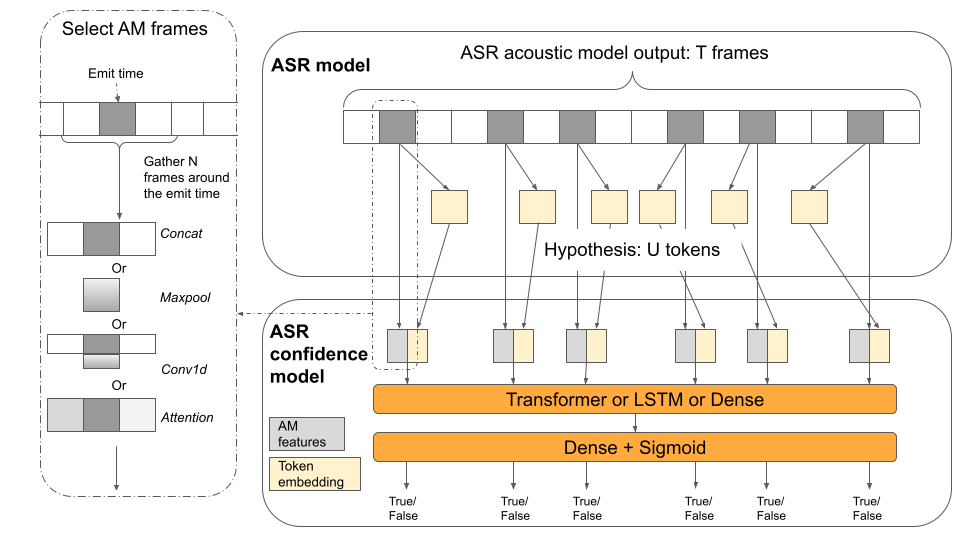}
    \caption{Confidence model architecture.}
    \label{fig:confidence_model}
\end{figure}

We propose a confidence model for RNN-T, which utilizes {\em acoustic and linguistic features} associated with output labels. For each hypothesized output label, we extract a latent acoustic representation from the ASR encoder outputs, stacking 3 frames of  left and right contexts at the time emission of the labels. The acoustic features are augmented with the embedding of the output labels (see Figure~\ref{fig:confidence_model}) and modeled with two transformer layers and 20 unit attention window. The proposed model incurs computation cost of $O(U)$ compared to $O(TU)$ in previous work for $T$ input frames and $U$ output labels~\cite{qiu2021learning}, allowing us to compute confidence over the segments even in streaming ASR applications.

While our best performing ASR systems operate at sub-word units such as graphemes and morphemes, users or downstream tasks require {\em confidence estimates at the world-level}. A straightforward solution involves accumulating the features at the word-level and predicting the estimates at the word-level. The downside of this approach is that the number of words are significantly lower than the number of graphemes by about a factor of 4.
We investigated an alternative approach where we force the model to predict all the graphemes to be incorrect even when only one of the graphemes in the associated word is incorrect and all the others are correct. The training examples were extracted after aligning ASR hypotheses with references. At inference time, the word level confidence score is estimated by averaging the scores of the corresponding sub-word units.

The performance of confidence models is primarily measured using normalized cross-entropy (NCE)~\cite{nce2017}. In practical applications, users or downstream applications may want to pick a confidence threshold to classify words as either correct or incorrect. Since current generation of ASR systems have substantially more correct predictions than incorrect ones, the most pertinent performance trade-off curve for picking a threshold is that with Negative Predictive Value (NPV) and True Negative Rate (TNR).

\vspace {0.2cm}
\noindent \textbf{Experimental Results}

\noindent The confidence models were trained on 6k held-out in addition to the training conversations and evaluated on the same ASR evaluation set. For reference, when the posterior probabilities of the (streaming morpheme) RNN-T model are used as confidence scores, the performance corresponds to an NCE of -0.40. Next, we attempted to regularize the confidence scores with the following techniques: 1) an RNN-T analog of label smoothing\footnote{Since RNN-T loss is calculated by forward-backward algorithm, regular label smoothing~\cite{muller2019does} cannot be directly applied. Instead, we sharpened the logits in the forward pass, and used un-sharpened logits in the backward pass.} (NCE = -0.14); 2) using temperature scaling~\cite{Guo2017} (NCE = -0.41); and 3) reducing language model overfitting by adding Gaussian noise to token embedding (NCE = -0.13), none of which helped.

\begin{table}[ht]
\centering
\begin{tabular}{|l|l|c|c|}
\hline
& Units & $\mathrm{NCE}$ & $\mathrm{AUC_{NPV~TNR}}$ \\ \hline
\multirow{2}{*}{non-streaming} & G	&	0.34	&	0.67 \\ \cline{2-4}
& M	&	0.38	&	0.73 \\ \hline
\multirow{2}{*}{streaming} & G	&	0.37	&	0.70 \\ \cline{2-4}
& M	&	0.30	&	0.58 \\ \hline
\end{tabular}
\caption{Confidence model performances for transformer RNN-T models with grapheme (G) and morpheme (M) as target units.}
\label{confidence:full_context}
\vspace{-.2in}
\end{table}

The proposed confidence model substantially improves the performance, with NCE scores consistently higher than $0.3$. The performance differences between the four systems were limited in terms of other metrics such as Expected Calibration Error (0.02 to 0.03), $\mathrm{AUC_{ROC}}$ (0.86 to 0.89), and $\mathrm{AUC_{PRC}}$ (0.95 to 0.97).

\section{Extracting Medical Information}
\label{sec:nlp}

\subsection{Task Description}
\label{nlp_corpus}
For developing the NLP models, providers and scribes annotated useful entities and attributes in the clinical conversations~\cite{shafran-etal-2020-medical}. The entities consisted of 64k symptoms, 100k medications, 42k conditions, 6k treatments and 38k diagnosis. The layman's description of symptoms were tagged using normalized labels. For example, "queasy" or "urge to vomit" were tagged with the clinical term "nausea". Medications were treated as an open set, and the catch all {\it Drug} label was applied to all direct and indirect references to drugs. This included specific and non-specific references, such as ``{\it Tylenol}'', ``{\it the pain medication}'', and ``{\it the medication}''. The attributes related to the medication entity that were annotated include {\it Prop:Frequency}, {\it Prop:Dosage}, {\it Prop:Mode}, and {\it Prop:Duration}. The conditions were categorized into -- Condition:Patient, Condition:Family History, and Condition:Other. 

The development and test sets were kept consistent with ASR corpus, described in Section~\ref{sec:asr_expts}. However, the training set is substantially smaller, with 5000 conversations, since labeling was labor-intensive.

\subsection{Modeling Architecture}
Linear-Chain Conditional Random Fields (CRF) are often used for named entity tagging~\cite{lafferty2001}. They work well when the label set is small, but are computationally expensive if multiple ontologies are modeled jointly as in our case. A hierarchical approach decomposes the problem by using a separate span extraction and attribute tagging layer~\cite{Nan-SAT}.

Sequence-to-Sequence with Attention models have also been used for tagging tasks~\cite{Kannan2018}. The target sequence is formed as a sequence of labels, not including \verb+<O>+ (BIO annotation). While this model does not suffer from large label spaces, the labels are not assigned to a specific location in the input text, making the output less interpretable.

To the best of our knowledge, RNN-Ts have not been used for named entity tagging tasks yet. They offer the same benefits as other sequence-to-sequence models and, additionally, specify the location of the labels in the input. The model scales easily for large label sets, and the decoder can capture dependencies between them, e.g. the prediction of {\em drug:tylenol} can benefit from an earlier prediction of {\em sym:pain} in the sequence. Unlike the conventional tagging scheme, the target sequence contains only the begin and the end symbols of each type of tag and the location is determined by emission time steps.

\subsection{Fixed Alignment Training of RNN-T}
Similar to other sequence-to-sequence models, RNN-Ts learn alignments, specifically by maximizing the marginal $P(Y|X)$, computed by summing over possible alignments, $P(Y|X) = \sum_{A} P(Y|X,A)$. The training effectively ignores the human annotated alignment information about the location of tags in the input, making the training task harder than it needs to be.

We propose a modification that leverages the human annotation better. We encode the position information as a sequence of {\em blank} and {\em label} steps that forms a valid path in the RNN-T alignment trellis and modify the objective function such that only this specific path is used, i.e. we maximize directly $P(Y|X,A_{h})$, where $A_{h}$ is the alignment from the human annotations. The model does not attempt to learn alignments anymore, instead they are given. It still models the joint dependencies between inputs and outputs and the inference is performed with Viterbi beam search like regular RNN-Ts.

\subsection{Experimental Results}
The main results are summarized in table~\ref{nlp:results}, where we compare a standard RNN-T with the fixed alignment model. The encoder is a transformer similar to the one used for ASR, and is pre-trained with a masked LM loss~\cite{MaskedLM,Bert} using all the 100k conversations. Note, our tagging model is the same as our ASR model, except for the fixed alignment loss.

We trained the models with the sequence length of about 50 words for regular RNN-T. Training regular RNN-T models with longer sequences leads to a substantial performance drop as the alignment learning problem becomes exacerbated, while the fixed alignment RNN-T becomes robust for long sequences in inference. Overall, the fixed alignment model gives improvements over the baseline RNN-T especially for the attributes, conditions, diagnostics and treatment. 

In a second set of experiments, we used the 95k unlabeled conversations for self-training, as reported in the last column of table~\ref{nlp:results}, we see a small improvement in symptoms, attributes, conditions and diagnostics. 

\begin{table}[t]
\centering
\begin{tabular}{|l|c|c|c|c|} \hline
Ontology      & RNN-T & RNN-T-FA & + Self-Train   \\ \hline
Overall                 & 0.72  &   0.75    &   0.76 \\ \hline
Symptoms                & 0.74  &   0.74    &   0.76 \\ \hline
Medication              & 0.90  &   0.90    &   0.90 \\ \hline
Attributes              & 0.58  &   0.66    &   0.67 \\ \hline
Condition               & 0.72  &   0.74    &   0.75 \\ \hline
Diagnostics             & 0.69  &   0.75    &   0.76 \\ \hline
Treatment               & 0.56  &   0.61    &   0.59 \\ \hline
Meds/Properties         & 0.69  &   0.69    &   0.69 \\ \hline
\end{tabular}
\caption{Named Entity Tagging Results (F-Measure). RNN-T-FA refers to the Fixed Alignment model.}
\label{nlp:results}
\vspace{-.3in}
\end{table}

\subsection{Error Analysis}
In previous work, we analyzed common errors observed in extracting clinical information~\cite{shafran-etal-2020-medical}. About $30\%$ of the residual errors were related to bad references. We notice similar instances in this model too. The model correctly tags the term "grass allergies", but the references misses this tag.\\[0.1cm]
{\em \footnotesize
    we know that you have SYM:IMMUNO:ALLERGIES grass allergies END:SYM because we did our DIAGS:LABS testing END:DIAGS
}
\\[0.1cm]
Another common error relates to recognizing only a portion of the relevant text. For example, the model only highlights {\em water} while the reference highlights the complete phrase {\em eyes would just water}\\[0.1cm]
{\em \footnotesize
    because I used to have this burning and my eyes would just SYM:EYES:WATEREYE water END:SYM uh-hum
}
\\[0.1cm]
Overall, these observations suggest that the automated metrics maybe underestimating the performance of the models.

\section{Conclusion}
In this paper, we have described three main components for extracting clinical information from medical conversations. Through empirical evaluation on real-world test set, we show that an RNN-T with a transformer encoder and grapheme units can provide better performance in a streaming (WER$\approx$20\%) scenario than LSTMs encoders or larger morpheme units. Our confidence model performs at 0.37 NCE which is comparable to other models in recent work on relatively simpler task with shorter utterances~\cite{qiu2021learning}. We demonstrate how a novel and relatively simpler RNN-T model with a fixed-alignment loss can improve performance over previous work that utilizes a tailored hierarchical model architecture~\cite{Nan-SAT}.

\section{Acknowledgements}
We are grateful for help and support from Hasim Sak, Anshuman Tripathi, Han Lu, Qian Zhang, Nan Du, Alex Greve, Ken Su, Roberto Santana, and Greg Corrado.

\newpage
\bibliographystyle{IEEEtran}
\bibliography{paper}

\end{document}